\documentclass[letterpaper, 10 pt, conference]{ieeeconf}  

\IEEEoverridecommandlockouts                              

\usepackage{graphics} 
\usepackage{amsmath} 

\usepackage{amsthm}
\usepackage{xcolor}

\usepackage{amsmath,amsfonts,amssymb}

\usepackage{url}            
\usepackage{booktabs}       
\usepackage{amsfonts}       
\usepackage{nicefrac}       
\usepackage{microtype}      
\usepackage{graphicx}
\usepackage{color}

\usepackage{bm}
\usepackage{cite} 

\usepackage{multirow}
\usepackage{multicol}
\usepackage[linesnumbered,ruled]{algorithm2e}
\graphicspath{{figure/}}
\usepackage{threeparttable}
\usepackage{adjustbox}
\usepackage{array}
\usepackage{dcolumn}

\usepackage{hyperref}       
\usepackage{cleveref}

\title{\LARGE \bf
HiCRISP: An LLM-based Hierarchical Closed-Loop Robotic Intelligent Self-Correction Planner
}

\author{Chenlin Ming$^{1}$, Jiacheng Lin$^{2}$, Pangkit Fong$^{1}$, Han Wang$^{3}$, Xiaoming Duan$^{1}$and Jianping He$^{1}$
\thanks{$^{1}$The Department of Automation, Shanghai Jiao Tong University, and Key Laboratory of System Control and Information Processing, Ministry of Education of China, Shanghai, China.
        Email: {\tt\small \{mcl2019011457, fpjgaoge, xduan, jphe\}@sjtu.edu.cn. }}%
    \thanks{$^{2}$ The Department of Computer Science, University of Illinois Urbana-Champaign, IL, USA. 
    Email: {\tt\small jl254@illinois.edu}.}
    \thanks{$^{3}$The Department of Engineering Science, University of Oxford, Oxford, UK. E-mails: {\tt\small han.wang@eng.ox.ac.uk}.}.
    \thanks{The website is available at \href{https://ming-bot.github.io/HiCRISP.github.io}{HiCRISP.github.io}.}
}

\begin{document}

\maketitle
\thispagestyle{empty}
\pagestyle{empty}

\begin{abstract}

The integration of Large Language Models (LLMs) into robotics has revolutionized human-robot interactions and autonomous task planning. However, these systems are often unable to self-correct during the task execution, which hinders their adaptability in dynamic real-world environments. To address this issue, we present an LLM-based \textbf{Hi}erarchical \textbf{C}losed-loop \textbf{R}obotic \textbf{I}ntelligent \textbf{S}elf-correction \textbf{P}lanner (HiCRISP), an innovative framework that enables robots to correct errors within individual steps during the task execution. HiCRISP actively monitors and adapts the task execution process, addressing both high-level planning and low-level action errors. Extensive benchmark experiments, encompassing virtual and real-world scenarios, showcase HiCRISP's exceptional performance, positioning it as a promising solution for robotic task planning with LLMs.


\end{abstract}

\section{Introduction}

Recently, the integration of LLMs into robotics has witnessed remarkable progress, empowering robots with natural language understanding and the ability to generate task plans directly from textual instructions \cite{singh2023progprompt, liang2023code,brohan2023can,jansen2020visually,li2022pre,patel2021mapping}. These LLMs, such as GPT-4 \cite{achiam2023gpt}, have demonstrated exceptional prowess in natural language understanding and generation, making them promising tools for enhancing human-robot interactions and autonomous task planning. Moreover, in contrast to conventional robotic systems, the integration of LLMs furnishes robots with the capability to perform a wide array of everyday tasks akin to those undertaken by humans~\cite{wu2023tidybot}. This enhancement has the potential to propel smart robot systems into novel frontiers.

While the integration of LLMs into robotics holds great promise. One limitation that persists in the current landscape of LLM-based robotics systems is their inability to self-correct during the execution of tasks. Specifically, when an LLM generates a task plan comprising multiple sequential steps from textual instructions, the robot may experience delays in plan execution and lack the ability to handle deviations resulting from factors such as environmental changes. These systems typically proceed with the predetermined plan, often resulting in task failures. This limitation hampers their adaptability in real-world environments where uncertainties and unforeseen obstacles are commonplace. 

To empower robotic systems with the capability to address unexpected failures, researchers have various solutions. ProgPrompt \cite{singh2023progprompt} employs assertion checking within the generated Python programs to ensure that actions continue to execute when confronted with errors. However, such rigid rule-based approaches usually fail to address unforeseen failures and unexpected situations. To develop more flexible methods, REFLECT \cite{liu2023reflect} and CLAIRify \cite{skreta2023errors} utilize the capabilities of LLMs to detect and correct robot failures by incorporating a failure feedback mechanism. Nonetheless, these methods primarily engage in error correction once long-horizon tasks have been completed, lacking the ability to rectify errors on a step-by-step basis during task execution. Consequently, their efficiency and effectiveness are limited. 

\begin{figure}
    \centering
    \includegraphics[width=\linewidth]{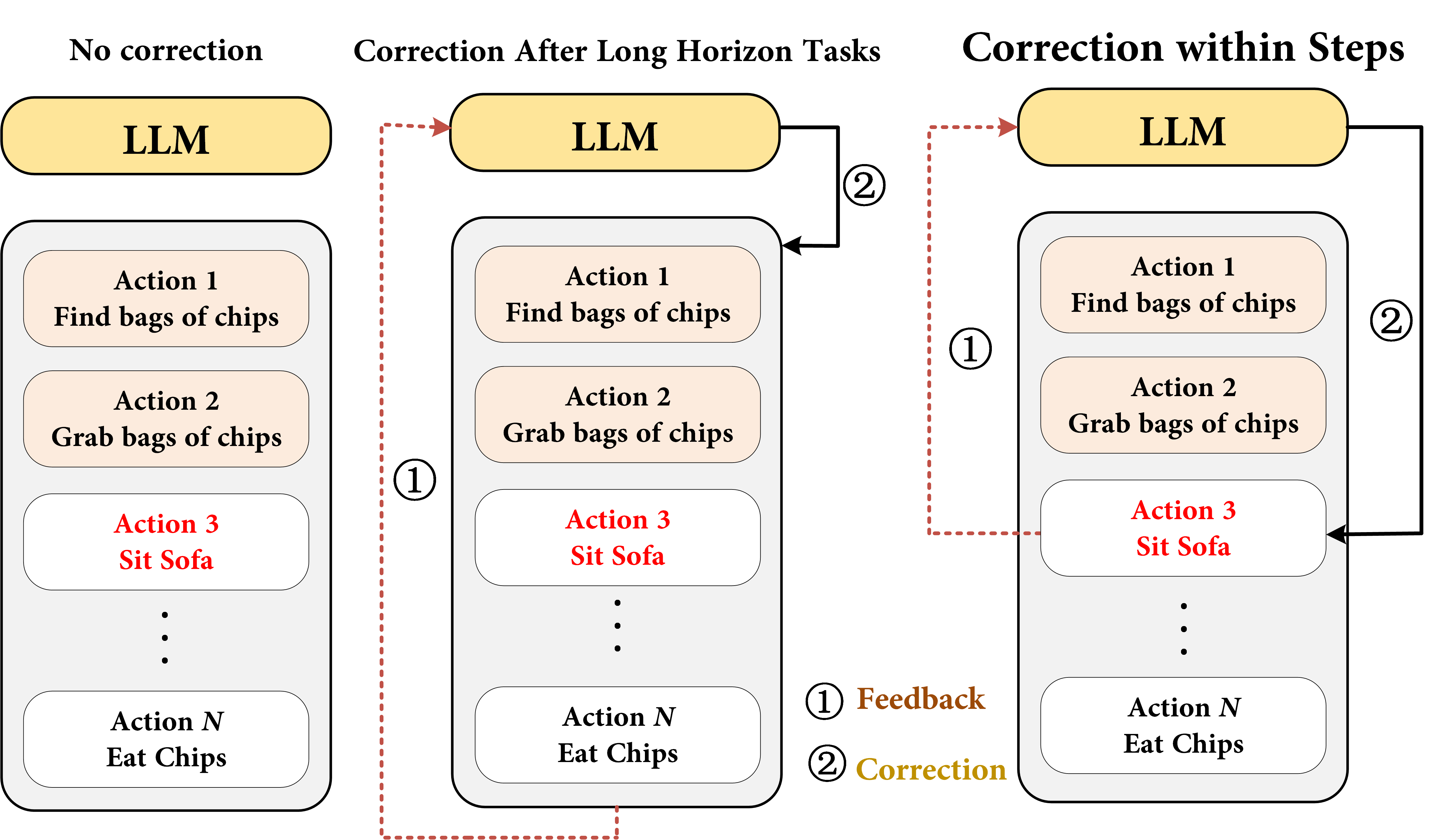}
    \caption{Illustration of LLM-based robotic systems: (1) without correction, (2) correction after task completion, and (3) correction within steps. An error occurs during the execution of Action 3. Our proposed HiCRISP belongs to the system that performs corrections within individual steps.}
    \label{fig:intro_motivation}
    \vspace{-1em}
\end{figure}

\begin{figure*}[t]
    \centering
    \includegraphics[width=1\linewidth]{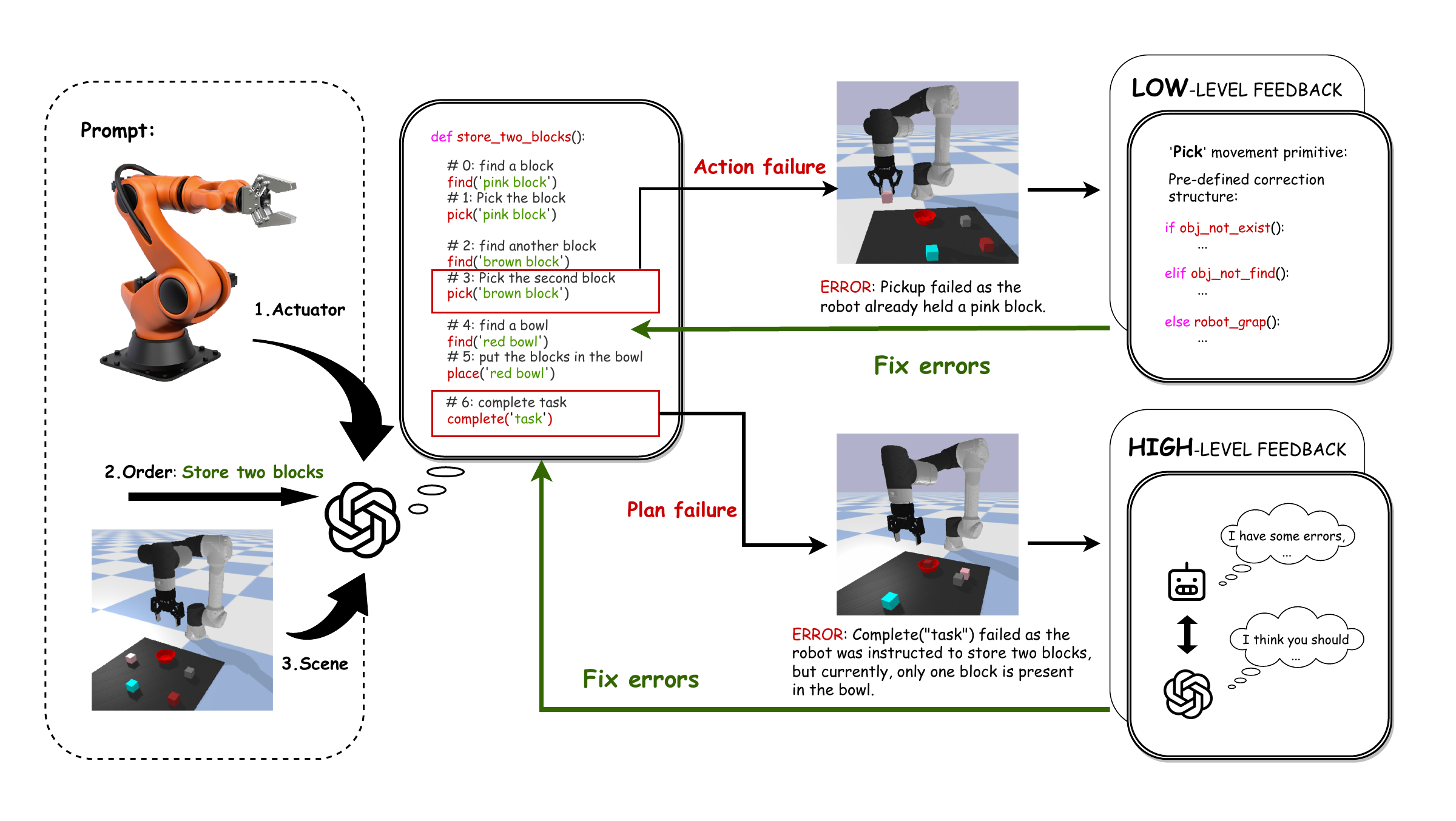}
    \caption{Overview of our proposed HiCRISP framework. The LLM dissects the user input into various actions using pertinent details from the actuator, scene, and other sources. We utilize predefined code to translate actions into movement primitives that are easily executable by the robot. Perception detects environment information and judges whether the system state changes. If a failure is detected, the system fixes plan failure and action failure through high-level feedback in \ref{subsec: high-level} and low-level feedback in \ref{subsec: low-level}, respectively.}
    \label{fig: System}
    \vspace{-1em}
\end{figure*}

To address the aforementioned issues and unlock the potential of LLM-based robotics, in this paper, we present our innovative \textbf{Hi}erarchical \textbf{C}losed-loop \textbf{R}obotic \textbf{I}ntelligent \textbf{S}elf-correction \textbf{P}lanner (HiCRISP). Our approach stands out by its capacity to address errors within individual steps during task execution, rather than waiting until long-horizon tasks are completed, shown in Fig. \ref{fig:intro_motivation}. By further incorporating hierarchical self-correction, our proposed HiCRISP can handle both high-level planning errors and low-level action errors, which promotes the success rate of tasks. Our main contributions can be summarized as:
\begin{itemize}
    \item We introduce a novel framework that integrates an automatic closed-loop self-correction for robot planning. This system enables robots to actively monitor and adjust their task execution.
    
    \item Our proposed system operates on a hierarchical structure, addressing both high-level and planning errors and low-level action errors. This hierarchical structure allows the robot to correct errors across different levels of closed-loop, thereby enhancing its overall robustness and adaptability.
    
    \item Through extensive experiments, we showcase the remarkable advantages of our proposed framework.
    Our results highlight the significant performance improvements achieved by integrating self-correction into the robotic planning process. We present evidence of our system's superiority in various benchmarks, underscoring its potential in future research.
\end{itemize}


\section{Related Works}

\noindent\textbf{Robot Task Planning with LLMs.} Integrating Large Language Models (LLMs) into robotic task planning has garnered significant attention in recent years. In the field of robotics research adding natural language, experts progressed in the areas of skill learning\cite{xiao2022robotic,ha2023scaling,shao2021concept2robot,yu2023language}, control performance\cite{ma2023liv,zitkovich2023rt,lin2023text2motion}, collaboration\cite{mandi2023roco}, human-robot interaction\cite{lynch2023interactive}, navigation\cite{shah2023lm}, etc. Further, researchers have explored the potential of LLMs to understand high-level instructions and generate low-level step-by-step actions that can be executed by agents directly from textual inputs \cite{brohan2023can, singh2023progprompt, liang2023code, huang2022language}. SayCan \cite{brohan2023can} employs LLMs to complete embodied tasks by combining pre-trained skills and learns value functions to choose contextually appropriate actions. Code as Policies (CaP) \cite{liang2023code} employs a structured code generation approach for robot control, enabling iterative incorporation of all necessary functionalities. VoxPoser \cite{huang2023voxposer} utilizes 3D Value Maps for robotic manipulation, employing code modules generated by LLMs to plan and control robots. This method allows for re-planning in case of task failures, enhancing the system's robustness. Inner Monologue \cite{huang2022inner} investigates how LLMs in embodied context can use natural language feedback to form an inner monologue, improving their processing and planning in robotic control scenarios. ProgPrompt \cite{singh2023progprompt} introduces a programmatic LLM prompt structure that enables functional plan generation in diverse robotic environments, accommodating various robot capabilities and tasks. The methods mentioned above demonstrate the significant potential of LLMs in the field of robot task planning.

\vspace{1em}
\noindent\textbf{Task Failure Correction with LLMs.} Despite the potential of integrating LLMs in robotics, robots face challenges in executing LLM-generated actions in some occasions. These challenges encompass cases where actions generated by LLMs cannot be performed due to their impracticality or unsuitability for the given context. Additionally, unexpected environmental changes can further hinder task execution by making the upcoming actions infeasible. In response to these failure situations, researchers explore methods to rectify such infeasible actions and address the disruptive impact of unforeseen factors on robotic task execution. REFLECT \cite{liu2023reflect} is a framework that employs LLMs to automatically detect and correct robot failures using past experiences. However, failure correction cannot commence until all processes have concluded, thus impeding the correction of errors occurring at a specific step of task execution. Similarly, DoReMi \cite{guo2023doremi} presents a framework that employs a set of error action constraints to resolve discrepancies between planning and execution. Nonetheless, it predominantly targets navigation tasks. CLAIRify \cite{skreta2023errors} leverages iterative prompting, integrates error feedback, and employs program verification to produce syntactically valid robot task plans from high-level natural language instructions. While it excels in error correction, CLAIRify specializes in generating long-horizon task plans and may not correct tasks on a step-by-step basis without completing all the programs.  Similarly, corrective re-prompting proposed in \cite{raman2022planning} enhances task plans by injecting pre-condition error information into LLMs but focuses on the prompting-based strategy rather than the complete robotic system framework.  In contrast, our proposed framework offers broader applicability by hierarchical self-correction in both long and short-horizon robotics task planning scenarios.

Beyond the field of robotics, research on failure correction with LLMs is conducted in various other domains, including agent-based systems and code generation. Reflexion \cite{shinn2023reflexion} is a framework that enhances language agents' learning in goal-driven tasks by utilizing linguistic feedback. In the code generation area, Self-Debugging \cite{chen2023teaching} teaches LLMs to debug the predicted program by few-shot demonstrations, even without any feedback on the code error messages. Compared with these methods, HiCRISP focuses on robot controlling and task planning by integrating closed-loop and hierarchical self-correction.

\section{Methodology}


In this section, we explain how we model the system which is composed mainly of an LLM and a robot. The LLM acts as the robot's planner, planning task sequence and determining executable action. The robot, as an anchor for interacting with the environment, executes commands from the LLM. Combined with perceptive information, we propose a framework named HiCRISP which can rectify itself by automatic feedback (Fig. \ref{fig: System}).

\subsection{Problem Description}

\subsubsection{Task Planning Formulation}
\label{subsec: problem}

We model the problem as a finite Markov decision process (MDP), typically represented as a tuple $\left \langle \mathcal{S}, \mathcal{A}, \mathcal{P}, \mathcal{R}\right \rangle$. $s \in \mathcal{S}$ represents the state information of the robot and environment. At each state, the robot chooses a semantic action $a(s) \in \mathcal{A}$. $r(s) \in \mathcal{R}$, accelerating preference learning, is provided by the perception part in the format of a text prompt. $\mathcal{P}_{ss^{'}}^a$ represents the probability of transitioning from the state $s$ to the state $s^{'}$ by executing the action $a$. 

In our framework, the LLM breaks down abstract high-level instructions into intermediate task states $\{s_i\}, i = 1, 2, \cdots, n$ and establishes a desired path using the sequence of semantic actions $\{a_i\}, i = 0, 1, \cdots, n$ from the initial state $s_0$. We define the success state as the termination state $s_{n + 1}$. The desired path is: 
\begin{align}
    s_0 \stackrel{a_0}{\longrightarrow} s_1 \stackrel{a_1}{\longrightarrow} s_2 \stackrel{a_2}{\longrightarrow} \cdots \stackrel{a_{n}}{\longrightarrow} s_{n + 1}.
\label{pro: desired path}
\end{align}

However, in the actual execution process of the robot, the progression from one state $s_i \in \mathcal{S}$ to the subsequent state $s_{i + 1} \in \mathcal{S}$, triggered by the execution of the semantic action $a_i \in \mathcal{A}$, often encounters hindrances or limitations that may prevent a seamless transition. That is, at the state $s_{i}$, the probability of transitioning to state $s_{i+1}$ is denoted as $p_{s_{i}s_{i + 1}}$ and the probability of transitioning to error states $s_{\text{error}}$ is $1 - p_{s_{i}s_{i + 1}}$. We assume that the probability of transitioning from $s_{\text{error}}$ to the $s_{i + 1}$ is greater than 0. Coupled with the MDP we construct being finite, we can conclude that all states in the desired path are reachable.

\begin{algorithm}[t]
    \caption{HiCRISP} \label{alg:a1}
    \KwIn{User input commands: task}
    \KwOut{Observation data from perception module;}
    $\{a_i\}_{i=0}^n = LLM(\mathcal{S}_0, \text{predefined prompt}, \text{task})$;\\
    \For{$a_i$ in $\{a_i\}_{i=0}^n$}{
        flag = predefined Error Checking ($a_i$);\\
        \uIf{not flag}{predefined correction process;}
        Execute $a_i$;
        Perception returns flag, info;\\
        \uIf{not flag}{
            Initialize stack;
            stack.push([$a_i$, info]);\\
            \For{$i = 1, \cdots, \text{threshold}$}{
                $\psi$ = stack.top.info;\\
                $a_{\text{correction}} = LLM(\mathcal{S}_{\text{error}}, \text{prompt}, \psi)$;\\
                Execute $a_{\text{correction}}$;\\
                Perception returns flag, info;\\
                \uIf{flag}{
                    \For{$stack.depth() > 0$}{
                        $a_{\text{try}} = $stack.top.action;\\
                        Execute $a_{\text{try}}$;\\
                        Perception returns flag, info;\\
                        \uIf{flag}{stack.pop;}
                        \uElse{
                        stack.update([$a_{\text{try}}$, info]);\\
                        break;\\
                        }
                    }
                }
                \uElse{
                    stack.push([$a_{\text{correction}}$, info]);\\
                }
            }
        }
    }
\end{algorithm}


\begin{figure*}[t]
    \centering
    \includegraphics[width=0.9\linewidth]{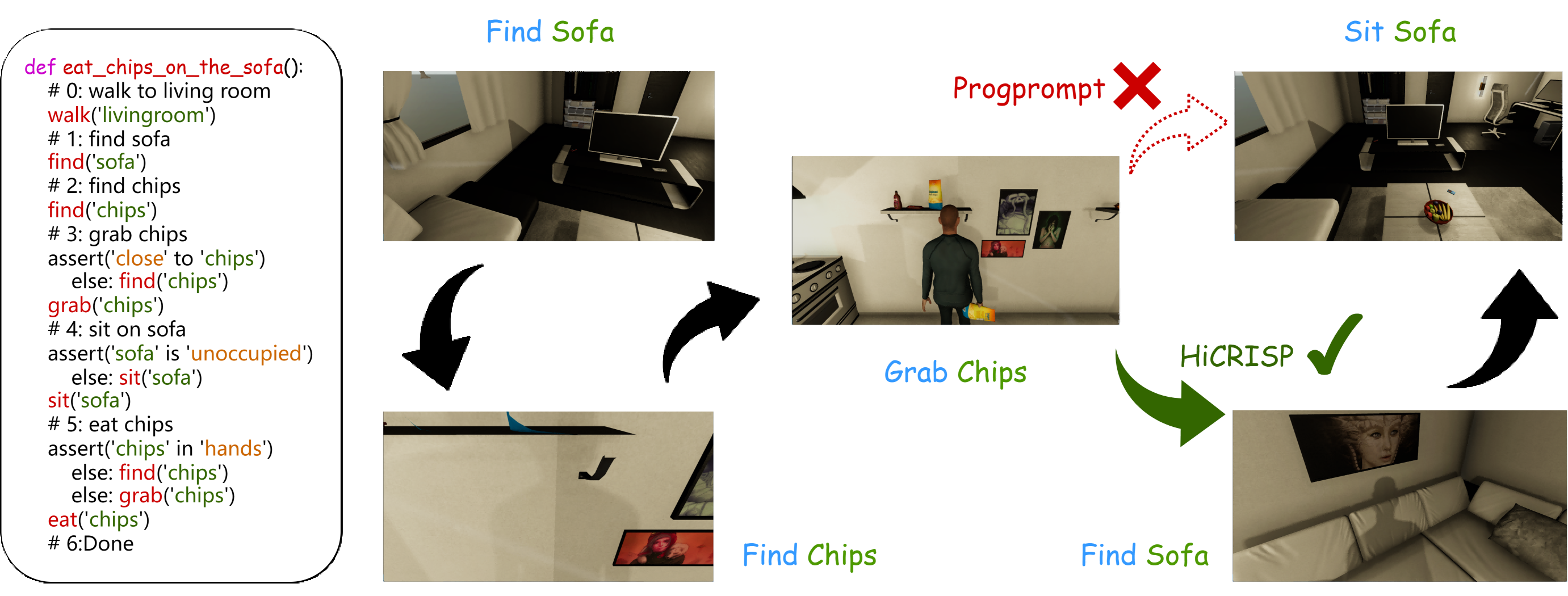}
    \caption{HiCRISP demonstrates superior performance in a VH simulator when compared to ProgPrompt\cite{singh2023progprompt}. In situations where the agent loses its position on the sofa and is unable to execute the command ``\texttt{Sit sofa}," HiCRISP exhibits the capability to learn from the error message provided by the simulator and take corrective action by executing the command ``\texttt{Find sofa}." This corrective action not only addresses the error but also guides the agent to reach the desired final state.}
    \label{fig: VHS}
    \vspace{-1em}
\end{figure*}

\subsubsection{Detecting Failures at High and Low Levels}

As our system requires controlling specific robots, we have defined movement primitives for robot motion. By using these predefined movement primitives in the semantic action $a_i$ generated by the LLM, we can control the robot to perform most of the predefined basic movements. In this architecture, we assume that errors can generally be categorized into two distinct types, with the possibility of both types of errors manifesting simultaneously.

\begin{enumerate}
    \item \textbf{High-Level Plan Failure}: In this scenario, all movement primitives in the $a_i$ generated by LLMs are executed successfully, yet the system fails to transition correctly to the intended subsequent state $s_{i + 1}$. As the system needs the LLM to rethink the plan, we categorize this type of failure as a high-level plan failure, thereby triggering our high-level feedback process.
    \item \textbf{Low-Level Action Failure}: Alternatively, there is another scenario where any one movement primitive in the $a_i$ generated by the LLM fails to execute successfully. For example, the executing movement primitive may lack certain prerequisites or conflict with the original constraints as the system advances. We classify this particular type of failure as a low-level action failure, leveraging our low-level feedback mechanism.
\end{enumerate}

\subsection{High-level Feedback}
\label{subsec: high-level}

Due to the limited information in the initial state $s_0$ and the inherent uncertainty of future environmental conditions, high-level state failure occurs sometimes, leading to the system failing to follow the desired path Eq. \eqref{pro: desired path}. More specifically, when the perception module detects that the direct successor of $s_i$ does not match $s_{i+1}$, our system initiates a closed-loop feedback loop. We consolidate language prompts by combining error messages and current status information, which is then fed back into the LLM. The LLM generates a rectified action $a_{\text{correction}}$. By executing $a_{\text{correction}}$, the current state is directed towards transitioning into $s_{\text{correction}}$, satisfying the condition: $s_{\text{correction}}\stackrel{a_i}{\longrightarrow} s_{i + 1}$ where the state $s_{i + 1}$ obtained by performing the same action $a_i$ is in the desired path. Mathematically, the entire process can be described as follows: when a transition system undergoes $s_i \stackrel{a_i}{\longrightarrow} s_{\text{error}}$, our framework adjusts it back to $s_{i + 1}$ in the desired path, represented as:
\begin{equation}
s_i \stackrel{a_i}{\longrightarrow} s_{\text{error}}\stackrel{a_{\text{correction}}^{1}} {\longrightarrow} s_{\text{correction}} \stackrel{a_{\text{correction}}^{2}} {\longrightarrow} \cdots \stackrel{a_i}{\longrightarrow} s_{i + 1}
\label{eq:highleq1}
\end{equation}

It should be noted that the $\stackrel{a_{\text{correction}}} {\longrightarrow}s_{\text{correction}}$ segment in Eq. \eqref{eq:highleq1} probably consists of multiple fragments to bridge state $s_{\text{error}}$ and $s_{i + 1}$. Moreover, state $s_{\text{error}}$ and $s_{\text{correction}}$ can be identical to $s_i$ and $s_{i + 1}$, respectively.

\subsection{Low-level Feedback}
\label{subsec: low-level}

As described in Section \ref{subsec: problem} and shown in Fig. \ref{fig: System}, we control the robot by translating the actions planned by the LLM into predetermined movement primitives. However, the real-world environment introduces numerous unforeseen deviations and noise interference, potentially perturbing the execution of these movement primitives and leading to task failures. To address the above issues, we propose a low-level feedback mechanism. To enhance efficiency, we pre-define some error-checking and correction structures within the movement primitives, enabling the robot's capability to rectify some low-level errors collected from error-checking structures without the need to consult the LLM. For unexpected errors that are undefined in movement primitives, we need the LLM to plan the correction action. As soon as the perception module detects error information, we utilize the feedback structure in Section \ref{subsec: feedback} to consult the LLM for resolutions.

\subsection{Error Correction Structure}
\label{subsec: feedback}


When the system faces $s_i \stackrel{a_{i}} {\longrightarrow} s_{\text{error}}$, the proceeding of the desired path Eq. \eqref{pro: desired path} should be halted. Persisting along the desired path without addressing the error is likely to fail in subsequent tasks. Thus, addressing the error first is crucial to establishing a solid foundation for long-term tasks. The correction feedback employs a stack-based approach adhering to the ``first in, last out" principle. Stack elements encompass executed actions and the associated error causes. 

The stack structure has a limited depth. Once the number of accumulated elements surpasses the predefined threshold, the feedback correction process is terminated. In this manner, while the error may not be fixed, the structure ensures that the system's state remains finite and prevents the system from getting stuck in a correction cycle that takes too long. The entire process can be summarized in Algorithm \ref{alg:a1}.
\begin{table}[htbp]
\newcommand{\tabincell}[2]{\begin{tabular}{@{}#1@{}}#2\end{tabular}}
\centering
\caption{ProgPrompt and HiCRISP performance on the VH test-time tasks.}
\resizebox{\linewidth}{!}{
\begin{tabular}{l|ccc|cc}
\toprule
& & Progrompt\cite{singh2023progprompt} & & \ \ \ HiCRISP & \\
Task description & $|A|$ & SR & Exec & SR & Exec\\
\midrule
Eat chips on the sofa & 6 & 0.12 $\pm$ 0.05 & \textbf{0.71 $\pm$ 0.00} & \textbf{0.33 $\pm$ 0.17} & \textbf{0.71 $\pm$ 0.00}\\
Put salmon in the fridge & 7 & 0.00 $\pm$ 0.00 & 0.80 $\pm$ 0.00 & 0.00 $\pm$ 0.00 & \textbf{0.90 $\pm$ 0.00}\\
Watch TV & 6 & 0.63 $\pm$ 0.00 & 0.89 $\pm$ 0.00 & \textbf{0.73 $\pm$ 0.10} & \textbf{1.00 $\pm$ 0.00}\\
Wash the plate & 10.5 & 0.00 $\pm$ 0.00 & 0.96 $\pm$ 0.00 & 0.00 $\pm$ 0.00 & \textbf{1.00 $\pm$ 0.00}\\
\tabincell{c}{Bring coffeepot and cup-\\cake to the coffee table} & 8 & 0.00 $\pm$ 0.00 & \textbf{0.71 $\pm$ 0.00} & 0.00 $\pm$ 0.00 & \textbf{0.71 $\pm$ 0.00}\\
Microwave salmon & 11.5 & 0.00 $\pm$ 0.00 & 0.80 $\pm$ 0.00 & 0.00 $\pm$ 0.00 & \textbf{0.87 $\pm$ 0.00}\\
Turn off light & 3 & 0.00 $\pm$ 0.00 & 0.67 $\pm$ 0.00 & 0.00 $\pm$ 0.00 & \textbf{1.00 $\pm$ 0.00}\\
Brush teeth & 11 & 0.17 $\pm$ 0.00 & 0.77 $\pm$ 0.00 & \textbf{0.18 $\pm$ 0.00} & \textbf{0.82 $\pm$ 0.00}\\
Throw away apple & 5 & 0.00 $\pm$ 0.00 & \textbf{1.00 $\pm$ 0.00} & 0.00 $\pm$ 0.00 & \textbf{1.00 $\pm$ 0.00}\\
Make toast & 10 & \textbf{0.16 $\pm$ 0.00} & \textbf{0.75 $\pm$ 0.00} & 0.15 $\pm$ 0.00 & \textbf{0.75 $\pm$ 0.00}\\
\bottomrule
\end{tabular}
}
\label{tab: virtualhome}
\vspace{-1em}
\end{table}

\begin{table*}[htbp]
\newcommand{\tabincell}[2]{\begin{tabular}{@{}#1@{}}#2\end{tabular}}
\centering
\caption{Ablation experiments with and w/o low-level feedback}
\resizebox{\linewidth}{!}{

\begin{tabular}{l|ccc|ccc}
\midrule
 &                    & with low-level feedback &                 &                    & without low-level feedback &                 \\
High semantic task description & $|\text{motions}|$ & $|\text{cues}|$         & Exec            & $|\text{motions}|$ & $|\text{cues}|$            & Exec            \\ \midrule
Store two blocks in the bowl                     & 10                 & 11                      & 1.00 $\pm$ 0.00 & 24                 & 50                         & 0.90 $\pm$ 0.10 \\
Put two blocks in one bowl step by step          & 7                  & 8                       & 1.00 $\pm$ 0.00 & 7                  & 15                         & 1.00 $\pm$ 0.00 \\
Put each block into each bowl                    & 8                  & 9                       & 1.00 $\pm$ 0.00 & 8                  & 17                         & 0.90 $\pm$ 1.00 \\
Put two blocks in a cool color bowl              & 16                 & 17                      & 1.00 $\pm$ 0.00 & 14                 & 30                         & 1.00 $\pm$ 0.00 \\
Put all warm color blocks in the green bowl      & 11                 & 5                       & 0.80 $\pm$ 0.20 & 18                 & 34                         & 0.75 $\pm$ 0.10 \\
Put blocks in bowls until only one bowl is empty & 19                 & 15                      & 0.70 $\pm$ 0.30 & 30                 & 50                         & 0.60$\pm$ 0.10  \\
Make only one bowl empty                         & 25                 & 19                      & 0.75 $\pm$ 0.00 & 30                 & 48                         & 0.50 $\pm$ 0.00 \\ \midrule
\end{tabular}
}
\label{tab: pybullet}
\vspace{-1em}
\end{table*}

\section{Simulation and Experiments}
To verify the effectiveness of our approach across diverse scenarios and objects, we conduct a series of experiments including both virtual simulations and real-world settings. This section provides detailed insights into high-level task planning (Section~\ref{subsec: virtualhome}), overall system operation (Section~\ref{subsec: simulator}), and real-world experiments (Section~\ref{subsec: realworld}). Our results demonstrate the framework's versatility, showcasing its applicability across various objects and scenarios. All systems can be conducted using GPT-3.5, with simulators built on ROS Noetic and Ubuntu 20.04 as the underlying infrastructure.
\subsection{Virtual Home Experiments}
\label{subsec: virtualhome}

First, we validate our HiCRISP on the Virtual Home Environment \cite{puig2018virtualhome} which contains a wealth of daily scenes and objects. Here, we focus on assessing the effectiveness of our error correction structure. In the simulation environment, the agent executes actions generated by the LLM planner, which occasionally results in task planning failure. Through a series of mission tests, compared with Progprompt \cite{singh2023progprompt}, our proposed framework effectively addresses these challenges. As illustrated in Fig. ~\ref{fig: VHS}, considering the command input ``\texttt{Eat chips on the sofa}", Progprompt follows the pre-generated subtasks but encounters failure while executing the ``\texttt{Sit sofa}" script. The essential reason for this failure is the wrong scheduling order in the task schedule. 
\begin{figure}[t]
    \centering
    \includegraphics[width=1\linewidth]{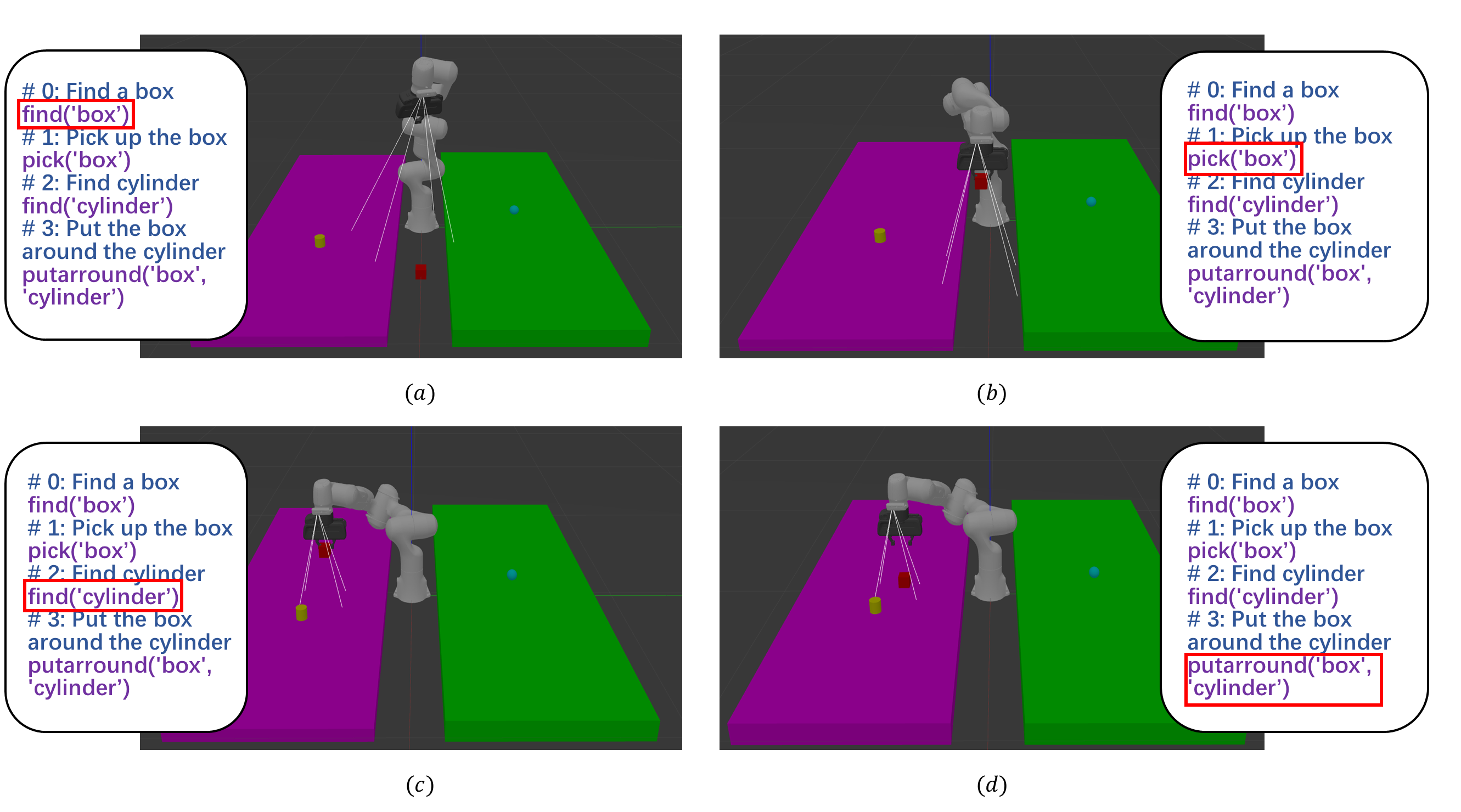}
    \caption{HiCRISP acts in Gazebo simulator. LLM breaks down the input: ``\texttt{Place a box around a cylinder}" into four steps: (1) locate a box; (2) pick up the box; (3) find a cylinder; (4) place the box around the cylinder.}
    \label{PandaSimulation}
    \vspace{-1em}
\end{figure}

We perform two iterations for various high-level tasks and record the result in Table \ref{tab: virtualhome}. $|A|$ shows the average length of semantic actions generated by GPT-3.5. $\emph{SR}$ represents success rate. $\emph{Exec}$ is the percentage of successfully executed semantic actions out of all actions attempted. Our method mainly focuses on improving the execution rate $\emph{Exec}$. Compared with the ``assert'' feedback in Progprompt \cite{singh2023progprompt}, HiCRISP executes more valid movements and achieves higher $\emph{SR}$ in some tasks. To ensure consistency in evaluation metrics, we adopt the method of calculating $\emph{SR}$ in Progprompt \cite{singh2023progprompt}, which involves predefining successful states to determine success. This evaluation approach overlooks the possibility of multiple ways to accomplish high-level commands, which contributes to the low $\emph{SR}$ in Table \ref{tab: virtualhome}.

\begin{figure*}[!htbp]
    \centering
    \includegraphics[width=0.9\linewidth]{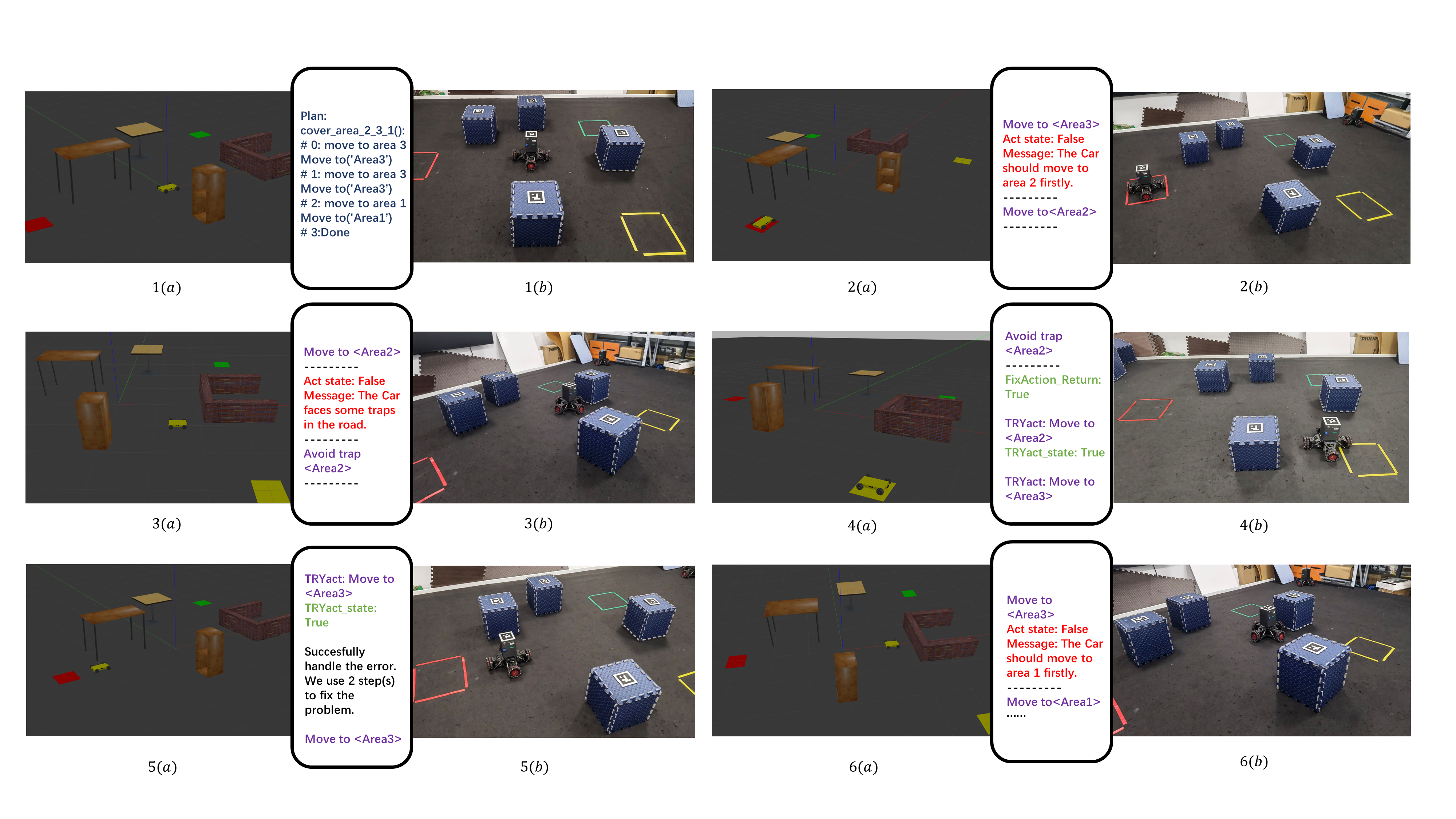}
    \caption{HiCRISP operates on both the Gazebo simulator and real-world AGV platform. We command the vehicle to sequentially approach specific landmarks according to a predetermined order. Action failure arises when obstacles obstruct the intended trajectory. Planning failure occurs if the vehicle erroneously navigates towards an incorrect landmark. HiCRISP addresses these failures by providing corresponding corrective actions and rectifying the issues.}
    \label{CarSimulation}
    \vspace{-2em}
\end{figure*}

\subsection{Robot Simulator Experiments}
\label{subsec: simulator}

To bridge the divide between NLP simulators and reality, we build HiCRISP in two real-world engine simulators, renowned for their realism. This part surpasses the use of natural language environments, as it enables us to closely simulate real-world scenarios.

\subsubsection{Gazebo Experiments}
To assess our framework's effectiveness, we perform pick-and-place tasks with the assistance of a Panda robot. We try to utilize the Visual Language Model (VLM) directly, like Gemini\cite{team2023gemini} and GPT-4\cite{achiam2023gpt}, as our perception module, but it fails to reason the correlated relationships in the scene. We attribute this to the fact that for the current VLM, understanding complex correlated relationships between objects in an image remains a significant challenge, at least not achievable through the input of a camera picture and simple prompts. However, as our contributions lie in correction feedback mechanisms, ultimately, we manually set perception parts to ensure the completeness of our system.

Considering potential deviations in location precision, we introduce an offset to the Panda robot's actuator. The offset intentionally increases the likelihood of not meeting the conditions when the Panda robot executes movement primitives. During simulated experiments shown in Fig. \ref{PandaSimulation}, when the perception issues an error message: ``The box is not close to the cylinder", the LLM promptly analyzes the received error message and current status information, then determines the appropriate action for self-correction.

\subsubsection{Bullet Experiments}

To demonstrate the universality of our framework, we build the table tasks following\cite{liang2023code}. We give the perception information manually to release the maximum ability of our feedback structure. To show our advantages in low-level feedback, we record the number of motion primitives that were generated $|\text{motions}|$ by the LLM for each task. We also record the number of cues $|\text{cues}|$ provided manually which is positively correlated with the calculation resource consumed. For comparison, we also built a similar system without low-level feedback that provides all error information to the LLM directly for feedback.

As shown in Table \ref{tab: pybullet}, most of the time, HiCRISP with low-level feedback fixes errors with fewer steps and receives less assistance from perception. Fewer steps mean shorter time and greater efficiency in handling errors which is crucial in real-life robot control. Requiring perception (VLM or human) fewer times means consuming fewer calculation resources which are now extremely expensive. 

\subsection{Real world experiment}
\label{subsec: realworld}

To further validate our framework, we conduct AGV experiments in both Gazebo simulations and real-world environments, utilizing a multi-robot test bed and indoor AGV systems\cite{ding2021robopheus}. As illustrated in Fig. \ref{CarSimulation}, the AGV is tasked with reaching different areas in a specified sequence. When the AGV inadvertently approaches incorrect areas, our predefined perception detects the errors and urges the LLM to re-plan, as depicted in Fig. \ref{CarSimulation}. Operating in an unfamiliar environment with limited perceptual range, unexpected obstacles along the planned path occasionally lead to action failures. Through low-level feedback, the movement primitives adapt to new constraints and generate revised trajectories. The results of these experiments demonstrate that the HiCRISP framework can be widely applied in robotics.

\section{Conculsion}
In this paper, we introduce HiCRISP, an LLM-based hierarchical robot control framework that restructures task planning and addresses erroneous messages, whether at the high level or low level, during robot missions via a feedback structure, thereby enhancing overall mission performance. By synergistically merging the feedback structure with the merits of LLM, HiCRISP exhibits the prowess to adeptly manage a wide array of tasks while possessing the ability to self-correct. Prospective areas of research could encompass augmenting the prowess of the perception module or delving into comprehensive and theoretically substantiated evaluations of system properties.







\bibliographystyle{IEEEtran}
\bibliography{ref}

\end{document}